\documentclass[pdflatex,sn-mathphys-num]{sn-jnl}
\usepackage{amsmath,amssymb,amsfonts}
\usepackage{booktabs,multirow,makecell,tabularx,array}
\usepackage{graphicx,adjustbox,float,xcolor,url}
\usepackage{microtype,placeins}
\newcommand{\method}{\textsc{HFCMA}}
\newcommand{\dgd}{\textsc{DGD}}
\newcommand{\ot}{\textsc{OT}}

\newcommand{\rsum}{\textsc{RSUM}}
\newcommand{\R}[1]{R@#1}
\newcommand{\tabfont}{\scriptsize}
\begin{document}
\title[Hierarchical Fine-Grained Cross-Modal Alignment]{Hierarchical Fine-Grained Cross-Modal Alignment via Dynamic Granularity Discovery and Optimal Transport}
\author[1]{\fnm{Pengfeng} \sur{Song}}\email{sungkites@163.com}
\author*[1]{\fnm{Shaozhong} \sur{Zhang}}\email{dlut\_z88@163.com}
\author[1]{\fnm{Pengfei} \sur{Shao}}
\affil[1]{\orgdiv{College of Information and Intelligent Engineering},\orgname{Zhejiang Wanli University},\orgaddress{\city{Ningbo},\postcode{315100},\state{Zhejiang},\country{China}}}

\abstract{Fine-grained image-text retrieval requires a model not only to assess global semantic consistency between an image and a text, but also to model visual evidence for local textual semantics, including objects, attributes, quantities, actions, and relations, so as to distinguish semantically similar candidates. Existing methods commonly use fixed image patches, detected regions, or text tokens as the basic matching units and then aggregate local interactions through max pooling, cross attention, or similarity accumulation. Such strategies remain limited in semantic unit construction and global matching consistency. This paper proposes a hierarchical fine-grained cross-modal alignment framework, termed HFCMA. The framework progressively transforms low-level patch and token representations into adaptive visual regions and textual phrases, and then performs structured region-phrase matching within a global retrieval scoring process. Specifically, HFCMA first adopts dynamic granularity discovery (DGD) to aggregate raw patches and tokens into semantically compact visual regions and textual phrases. It then formulates region-phrase matching as an entropy-regularized optimal transport problem, thereby estimating a structured alignment matrix under global marginal constraints. To reduce background noise and local semantic drift, HFCMA further introduces background debiasing, textual importance weighting, and prototype consistency regularization. Experiments on Flickr30K and MS-COCO show that \method{} effectively combines dynamic semantic unit construction with \ot{}-based structured matching under a dual-encoder retrieval framework. The resulting fine-grained score provides structured local alignment information for candidate ranking and improves the discrimination of subtle semantic differences.}
\maketitle

\section{Introduction}

Large-scale image-text contrastive learning has substantially advanced visual-language representation learning. CLIP and ALIGN demonstrate that global alignment over massive image-text pairs can learn transferable multimodal representations \cite{ref1,ref2}. Subsequent work, including ViLBERT, LXMERT, and VL-BERT, improves cross-modal interaction through fusion-style pre-training \cite{ref3,ref4,ref5}. OSCAR, ALBEF, and X-VLM further introduce object semantics, momentum distillation, and multi-grained visual concepts to strengthen image-text alignment \cite{ref6,ref7,ref8}. Meanwhile, ViLT, BLIP, BLIP-2, CoCa, and SigLIP extend visual-language pre-training from region-free modeling, generative learning, frozen-encoder bridging, and loss design \cite{ref9,ref10,ref11,ref12,ref13}. These advances have moved image-text retrieval from handcrafted features and region matching toward scalable pre-training frameworks.

Despite this progress, fine-grained image-text retrieval remains a weak aspect of current visual-language models. A global vector similarity is efficient for large-scale retrieval, but it is often insufficient when semantically similar candidates must be distinguished. For example, two images may contain the same object categories but differ in number, color, action, or spatial relation. A sentence may also describe only part of a visual scene rather than the whole image. In these cases, a retrieval model needs more reliable cross-modal correspondences at the levels of regions, attributes, and relations. Prior methods such as SCAN, VSR, UNITER, FILIP, SPARC, NAAF, CHAN, MSRM, CORA, and LAPS show that local interaction is important for improving the detailed discrimination of image-text retrieval \cite{ref14,ref15,ref16,ref17,ref18,ref19,ref20,ref21,ref22,ref23}.

Existing fine-grained approaches often build local interaction over backbone-produced patches, detector-based regions, tokens, words, or language-guided selected patches, and then design attention, graph reasoning, or similarity aggregation strategies for cross-modal matching. Although these designs improve local discrimination, the construction of matching units is still frequently constrained by fixed tokenization, external detectors, or predefined selection rules. As a result, the semantic granularity of local units may not be fully adaptive to each image-text pair. However, in real image-text pairs, a phrase may correspond to multiple adjacent patches, and a visual region may simultaneously carry object, attribute, and relation cues. If the matching units are semantically fragmented, subsequent local interaction can only partially remedy the mismatch. Recent studies on GroupViT, Slot Attention, TokenLearner, and Token Merging suggest that visual or textual representations can be adaptively organized into more compact semantic units \cite{ref29,ref30,ref31,ref32}. This provides a different perspective for fine-grained retrieval: the matching granularity should not be treated merely as a by-product of the backbone, but should be explicitly modeled.

Local matching also requires global consistency. Max pooling and cross attention often select the most similar local counterpart independently for each element, which may overlook the joint constraint between a set of visual regions and a set of textual phrases. Optimal transport (\ot) provides a principled formulation for structured set matching: given marginal distributions, all local correspondences are optimized jointly rather than independently \cite{ref33,ref34}. Although \ot{} has been used in distribution alignment, word mover distance, domain adaptation, and vision-language prompt learning \cite{ref35,ref36,ref37,ref38,ref39}, how to combine dynamic semantic unit discovery with \ot{} matching for fine-grained image-text retrieval remains under-explored.

Motivated by these observations, this paper proposes \method{}, a hierarchical fine-grained cross-modal alignment framework. \method{} addresses two questions: how to adaptively construct suitable visual regions and textual phrases for each image-text sample, and how to establish globally consistent region-phrase matching among these dynamic semantic units. \method{} adopts a frozen CLIP-style dual encoder as the feature backbone. It first uses a dynamic granularity discovery module to organize patches and tokens into semantically more complete local units. It then solves a structured transport matrix with entropy-regularized \ot{} and combines the resulting fine-grained score with the global similarity for candidate ranking.

The main contributions are summarized as follows.

\begin{itemize}
    \item We propose a dynamic granularity discovery mechanism for image-text retrieval, which bridges low-level patch/token representations and higher-level region-phrase semantic units, transforming fixed local matching into sample-adaptive region-phrase alignment.

    \item We formulate fine-grained cross-modal matching as entropy-regularized optimal transport over the discovered semantic units, enabling region-phrase correspondences to be jointly optimized under global marginal constraints.

    \item We introduce textual importance weighting, background debiasing, and prototype consistency regularization to improve the stability and discriminability of local semantic units.

    \item We conduct comprehensive experiments covering standard retrieval, module ablation, matching mechanism comparison, parameter sensitivity, inference efficiency, cross-dataset transfer, and visualization, showing both the effectiveness and the applicable boundary of HFCMA.
\end{itemize}

\section{Related Work}

\subsection{Visual-Language Pre-Training and Image-Text Retrieval}

Visual-language pre-training has evolved from fusion-based interaction to dual-encoder contrastive learning. ViLBERT, LXMERT, and VL-BERT jointly encode visual regions and text with cross-modal Transformers and achieve early progress on vision-language tasks \cite{ref3,ref4,ref5}. OSCAR, UNITER, ALBEF, and X-VLM improve cross-modal semantics through object tags, alignment objectives, momentum distillation, and multi-grained visual concepts \cite{ref6,ref7,ref8,ref17}. These methods provide strong interaction ability, but their pairwise fusion cost is high when many candidates must be evaluated.

CLIP and ALIGN encode images and texts into global vectors, making large-scale retrieval efficient through offline indexing and fast inner-product search \cite{ref1,ref2}. Later models, including BLIP, BLIP-2, CoCa, and SigLIP, strengthen visual-language foundation models through generative pre-training, frozen encoder bridging, contrastive-generative modeling, and sigmoid loss \cite{ref10,ref11,ref12,ref13}. However, the efficiency of dual encoders comes with a compression of local details. Global embeddings may not preserve object position, attribute modification, and relational cues completely. Therefore, this work focuses on enhancing fine-grained alignment while keeping the efficiency advantage of dual-encoder retrieval.

\subsection{Fine-Grained Cross-Modal Alignment}

Fine-grained image-text matching usually builds local interaction at region-word, patch-token, or phrase-region levels. SCAN models region-word correspondence with stacked cross attention \cite{ref14}. VSR and GraDual exploit visual or cross-modal graph reasoning for local relation modeling \cite{ref15,ref40}. UNITER introduces word-region alignment objectives \cite{ref17}, while FILIP pushes alignment down to the token level \cite{ref16}. Recent methods further study negative-aware attention, hard alignment, multilateral semantic relations, compositional relation modeling, and language-guided patch slimming, such as NAAF, CHAN, MSRM, CORA, and LAPS \cite{ref19,ref20,ref21,ref22,ref23}. Related studies also explore context-aware attention, graph matching, improved visual semantic reasoning, causal attention, and granularity-aware uncertain region modeling for fine-grained retrieval \cite{ref24,ref25,ref26,ref27,ref28}.

These studies demonstrate that fine-grained cues are important for distinguishing semantically similar image-text samples. Nevertheless, most methods still regard local matching units as fixed inputs and mainly focus on how to aggregate local similarities. Fine-grained alignment should also answer what units should be matched. If patches or tokens do not form complete semantic units, local interaction may amplify accidental high similarities among fragmented elements. Different from these methods, \method{} treats local semantic unit construction as an explicit part of retrieval-oriented matching. Instead of only improving the aggregation of local similarities, it first discovers sample-adaptive region and phrase units and then aligns them through a globally constrained transport plan.

\subsection{Dynamic Semantic Grouping and Adaptive Representation}

Dynamic semantic grouping aims to form compact and semantically coherent representations from low-level features. Slot Attention learns object-centric representations through competitive soft assignment \cite{ref30}. Set Transformer and Perceiver show that attention-based architectures can process variable-length sets and compress high-dimensional inputs \cite{ref41,ref42,ref43}. DETR decodes structured detection results with object queries \cite{ref44}. GroupViT demonstrates that visual patches can gradually form semantic groups under text supervision \cite{ref29}. TokenLearner and Token Merging reduce redundant representations through adaptive token selection or merging \cite{ref31,ref32}.

Most of these methods are designed for detection, segmentation, video understanding, or efficient inference rather than image-text retrieval. These studies motivate adaptive semantic organization, but they are not designed to optimize dynamically grouped units for bidirectional image-text retrieval. \method{} adapts this idea to retrieval by learning adaptive visual and textual semantic units under image-text supervision and aligning them with \ot{}-based structured matching.

\subsection{Optimal Transport and Structured Matching}

Optimal transport minimizes the overall transportation cost between sets or distributions and can jointly optimize a matching matrix under marginal constraints. The entropy-regularized formulation and Sinkhorn iteration introduced by Cuturi make \ot{} efficient and differentiable for deep learning \cite{ref33}. From Earth Mover's Distance to Word Mover's Distance and domain adaptation, \ot{} has become a useful tool for structured alignment \cite{ref36,ref37,ref38,ref39}. In visual-language learning, PLOT applies \ot{} to prompt learning for cross-modal transfer alignment \cite{ref35}.

However, existing \ot{}-based methods are usually applied over fixed-granularity features, where the transport objects are predefined patches, tokens, or prompts. \method{} places dynamic granularity discovery before \ot{}, so that the transport matrix acts on more compact visual regions and textual phrases. This design improves both the semantic quality of matching units and the global consistency of local alignment.

\section{Method}

\subsection{Problem Definition and Framework}

\begin{figure}[!t]
    \centering
    \includegraphics[width=0.98\textwidth]{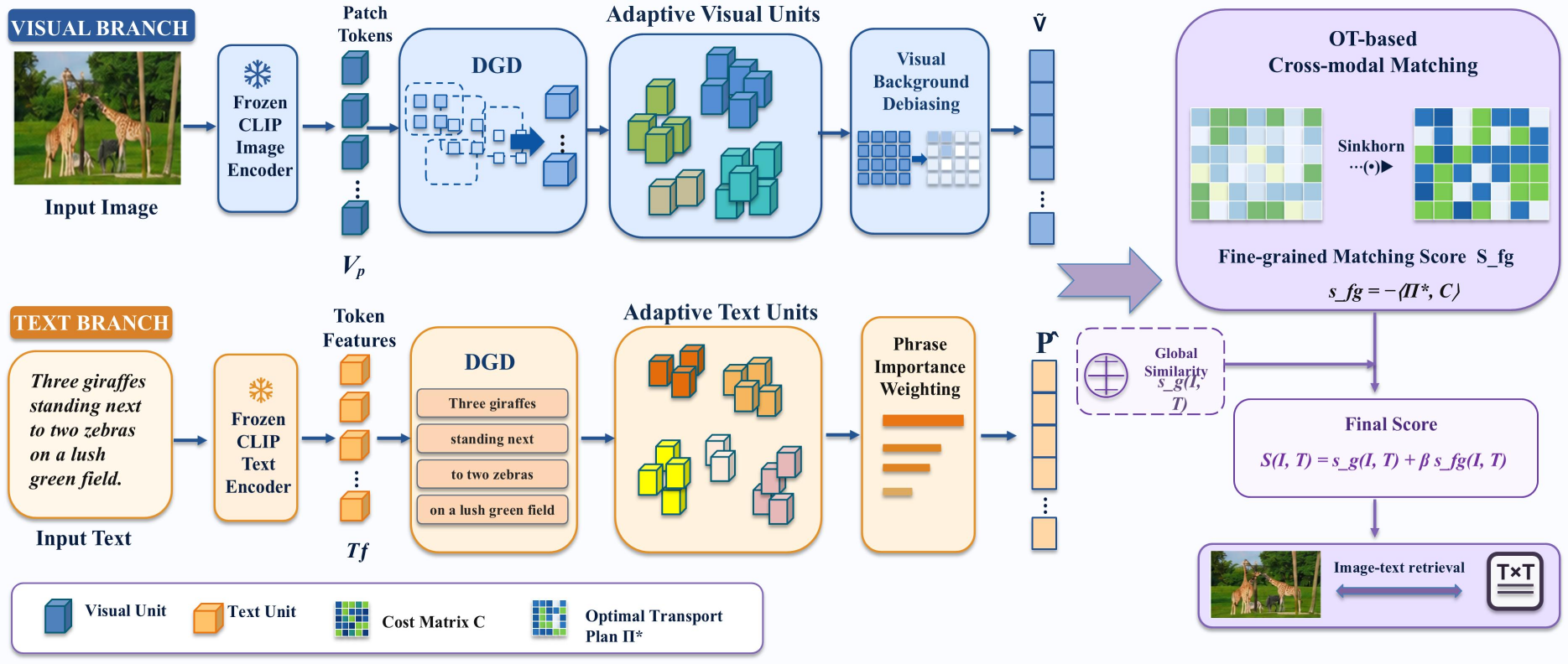}
    \caption{Overall framework of \method. A frozen CLIP-style dual encoder extracts visual patch and textual token features. Dynamic granularity discovery organizes them into adaptive visual and textual semantic units, followed by visual background debiasing and phrase importance weighting. The refined units are matched by entropy-regularized \ot{}, and the resulting fine-grained score is combined with global CLIP similarity for final image-text retrieval.}
    \label{fig:overall}
\end{figure}

Given an image $I$ and a text $T$, image-text retrieval learns a matching function $S(I,T)$ such that true image-text pairs receive higher scores than mismatched pairs. A frozen image encoder $E_v$ and a frozen text encoder $E_t$ produce global features and low-level local features:
\begin{equation}
\mathbf{g}^v, \mathbf{X}^v = E_v(I), \qquad
\mathbf{g}^t, \mathbf{X}^t = E_t(T),
\end{equation}
where $\mathbf{X}^v=\{\mathbf{x}_i^v\}_{i=1}^{N_v}$ denotes patch-level visual features and $\mathbf{X}^t=\{\mathbf{x}_j^t\}_{j=1}^{N_t}$ denotes token-level textual features. Instead of directly matching $\mathbf{X}^v$ and $\mathbf{X}^t$, \method{} first organizes them into semantically more complete visual regions and textual phrases, and then solves a structured cross-modal matching problem. The CLIP-style image and text encoders are kept frozen throughout training. \method{} only learns lightweight projection layers, dynamic granularity discovery modules, prototype parameters, and score fusion parameters. This preserves the efficiency of dual-encoder retrieval while adding structured local evidence for candidate ranking. In \method{}, the hierarchical design refers to this progressive pipeline from low-level patch/token representations, to adaptive region-phrase semantic units, and then to structured \ot{} matching combined with global retrieval scoring. Fig.~\ref{fig:overall} shows the overall framework.

The global similarity is computed as
\begin{equation}
S_g(I,T)=\cos(\mathbf{g}^v,\mathbf{g}^t).
\end{equation}
The fine-grained module produces visual semantic units $\mathbf{V}=\{\mathbf{v}_k\}_{k=1}^{K_v}$ and textual semantic units $\mathbf{P}=\{\mathbf{p}_l\}_{l=1}^{K_t}$, where $K_v$ and $K_t$ are the effective numbers of activated units. The final matching score is
\begin{equation}
S(I,T)=S_g(I,T)+\beta S_{fg}(I,T),
\end{equation}
where $S_{fg}$ is the \ot{}-based fine-grained score and $\beta$ controls the contribution of local alignment.

\subsection{Dynamic Granularity Discovery}

\begin{figure}[!t]
    \centering
    \includegraphics[width=0.98\textwidth]{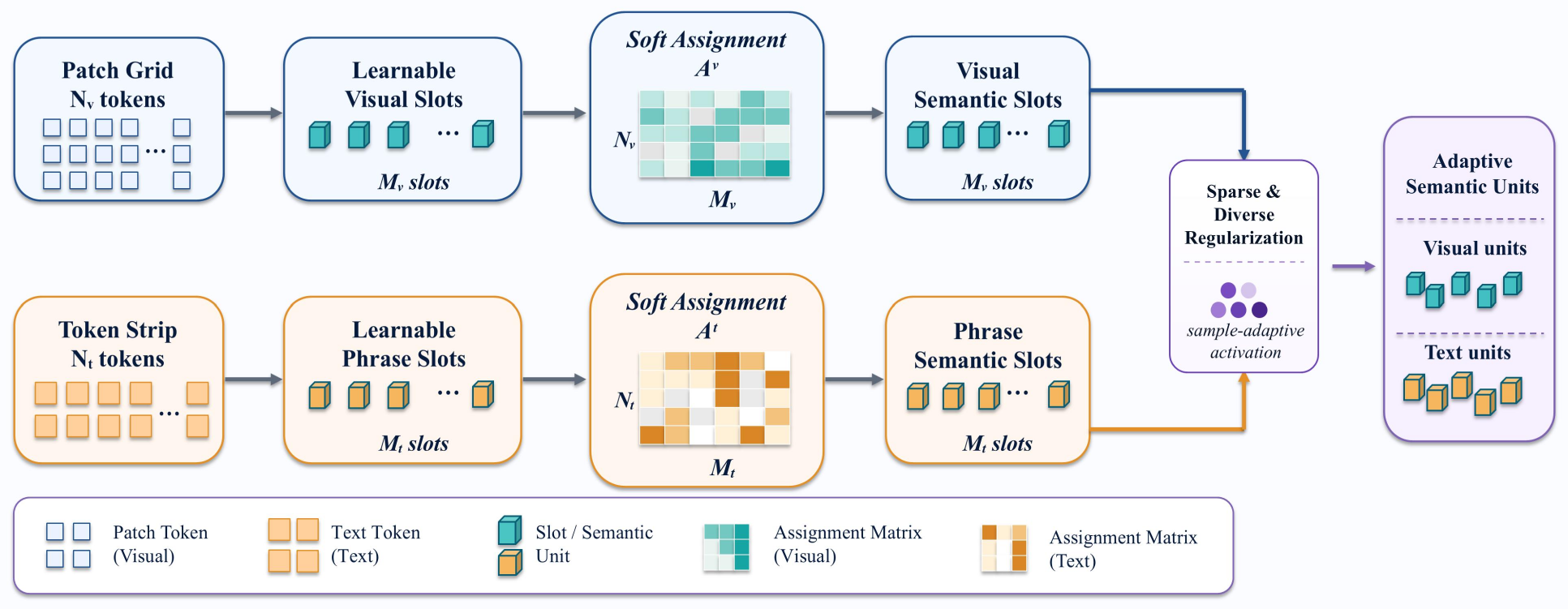}
    \caption{Illustration of dynamic granularity discovery. Patch and token sequences are softly assigned to learnable visual and phrase slots, producing compact semantic units without requiring bounding-box or phrase-boundary annotations. Sparse and diverse regularization encourages sample-adaptive slot activation and reduces redundant local units.}
    \label{fig:dgd}
\end{figure}

The goal of dynamic granularity discovery is to convert fixed patch/token sequences into semantic units closer to visual regions and textual phrases. For the visual branch, the model introduces $M_v$ learnable visual slots $\mathbf{Q}^v=\{\mathbf{q}_k^v\}_{k=1}^{M_v}$. The soft assignment between patch $i$ and visual slot $k$ is computed by
\begin{equation}
A_{ik}^v=\operatorname{softmax}_{k}
\left(
\frac{\phi(\mathbf{x}_i^v)^\top \psi(\mathbf{q}_k^v)}{\tau_v}
\right),
\end{equation}
where $\phi(\cdot)$ and $\psi(\cdot)$ are linear projections and $\tau_v$ is the assignment temperature. The $k$-th visual semantic unit is obtained by weighted aggregation:
\begin{equation}
\mathbf{v}_k=
\frac{\sum_{i=1}^{N_v} A_{ik}^v \mathbf{x}_i^v}
{\sum_{i=1}^{N_v} A_{ik}^v+\epsilon_0}.
\end{equation}
The text branch adopts the same mechanism with textual slots $\mathbf{Q}^t=\{\mathbf{q}_l^t\}_{l=1}^{M_t}$ and produces phrase units $\mathbf{P}=\{\mathbf{p}_l\}_{l=1}^{K_t}$. Compared with individual tokens, phrase units can better represent entity combinations, attribute modifiers, and relational fragments, such as ``two zebras'', ``standing next to'', or ``on a green field''. Fig.~\ref{fig:dgd} illustrates the conversion from fixed-granularity inputs to dynamic semantic units.

To control the number of effective slots and reduce repeated semantic responses, a lightweight granularity regularizer is used:
\begin{equation}
\mathcal{L}_{gran}
= \mathcal{L}_{cmp}
+ \lambda_s \mathcal{L}_{sparse}
+ \lambda_d \mathcal{L}_{div}.
\end{equation}
For modality $m\in\{v,t\}$, let $\mathbf{z}_i^m$ denote an input local feature, $\mathbf{u}_k^m$ denote a discovered semantic unit, and $A_{ik}^m$ denote the corresponding soft assignment. The compactness term is written as
\begin{equation}
\mathcal{L}_{cmp}
=\sum_m \frac{1}{N_m}\sum_{i=1}^{N_m}\sum_{k=1}^{M_m}
A_{ik}^m\left\|\mathbf{z}_i^m-\mathbf{u}_k^m\right\|_2^2 .
\end{equation}
The slot activation is $\rho_k^m=\frac{1}{N_m}\sum_i A_{ik}^m$. The sparsity term penalizes high activation entropy,
\begin{equation}
\mathcal{L}_{sparse}
=-\sum_m\sum_{k=1}^{M_m}\rho_k^m\log(\rho_k^m+\epsilon_0),
\end{equation}
so that only content-relevant slots remain active. The diversity term reduces repeated slot responses:
\begin{equation}
\mathcal{L}_{div}
=\sum_m \frac{1}{M_m(M_m-1)}
\sum_{k\ne k'}\max\bigl(0,\cos(\mathbf{u}_k^m,\mathbf{u}_{k'}^m)-\eta\bigr).
\end{equation}
Together, these terms encourage compact, selective, and non-redundant semantic units. They serve the retrieval objective and do not require a slot to correspond exactly to a human-annotated region or phrase boundary.

\subsection{\ot{}-based Structured Region-Phrase Matching}

After obtaining visual units $\mathbf{V}$ and phrase units $\mathbf{P}$, fine-grained matching is formulated as structured alignment between two semantic sets. The region-phrase cost matrix is defined by cosine distance:
\begin{equation}
C_{kl}=1-\cos(\mathbf{v}_k,\mathbf{p}_l).
\end{equation}
\method{} then solves an entropy-regularized \ot{} problem:
\begin{equation}
\Pi^{*}
= \arg\min_{\Pi \in \mathcal{U}(\mathbf{a},\mathbf{b})}
\langle \Pi,\mathbf{C}\rangle
+ \varepsilon \sum_{k,l}\Pi_{kl}(\log \Pi_{kl}-1),
\end{equation}
where $\Pi_{kl}$ denotes the soft matching strength between visual unit $k$ and phrase unit $l$, $\mathbf{a}$ and $\mathbf{b}$ are marginal distributions, and $\varepsilon$ controls the smoothness of the transport plan. The base setting normalizes the marginals over active semantic units, and the textual side can be further reweighted by phrase importance as described in Sec.~3.4. The problem is efficiently solved by Sinkhorn iteration, which alternately rescales rows and columns of the transport matrix to satisfy the marginal constraints.

\begin{figure}[!t]
    \centering
    \includegraphics[width=0.96\textwidth]{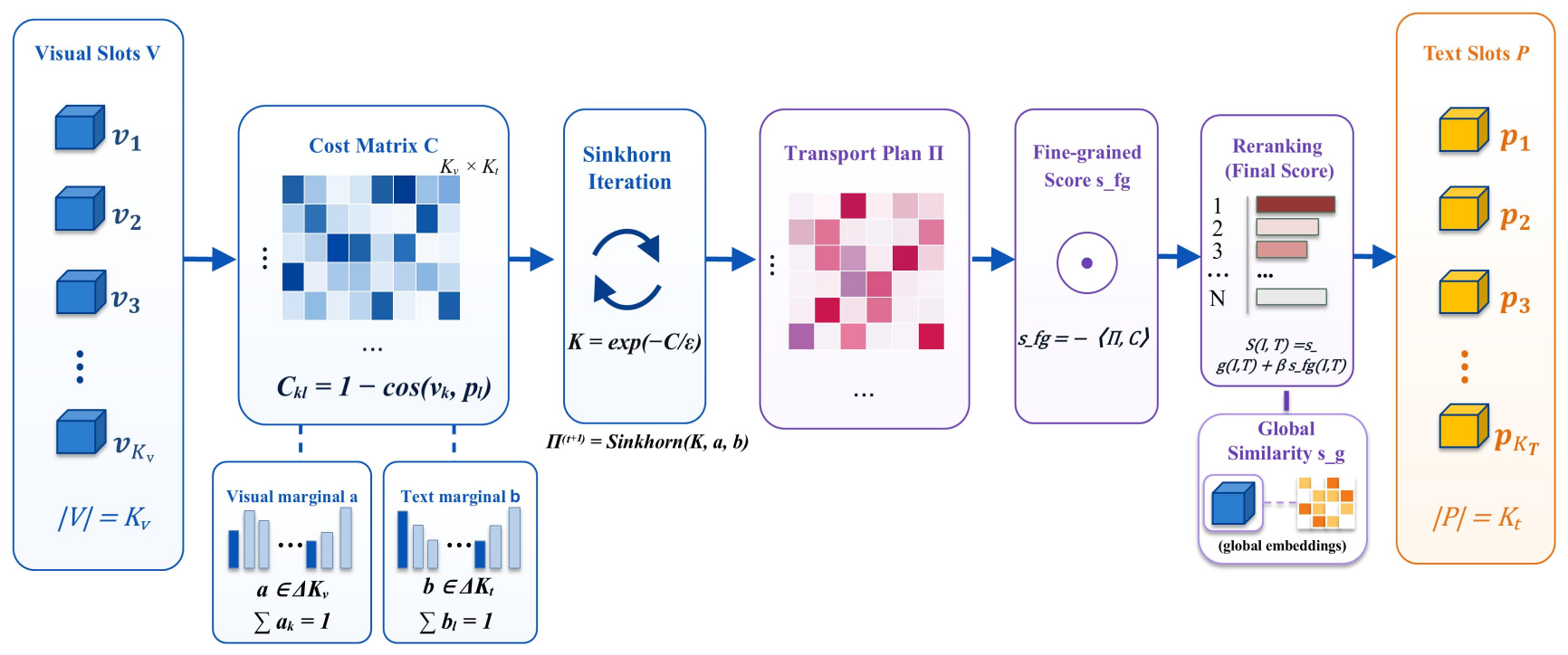}
    \caption{\ot{}-based structured matching between discovered visual and textual slots. \method{} builds a cross-modal cost matrix, estimates a transport plan with Sinkhorn iteration under visual and textual marginal constraints, and converts the transport cost into a fine-grained score for reranking together with the global similarity.}
    \label{fig:ot}
\end{figure}

The fine-grained score is defined as the negative transport cost:
\begin{equation}
S_{fg}(I,T)=-\sum_{k=1}^{K_v}\sum_{l=1}^{K_t}
\Pi^{*}_{kl} C_{kl}.
\end{equation}
Fig.~\ref{fig:ot} shows the structured matching process. Unlike hard assignment, \ot{} preserves uncertainty in local correspondence. Unlike max pooling, \ot{} reduces the risk that multiple phrases compete for the same highly similar visual region. It is therefore more suitable for many-to-many fine-grained matching.

The main computational cost comes from Sinkhorn matching for candidate pairs. If the number of visual slots is $K_v$, the number of text slots is $K_t$, and the number of Sinkhorn iterations is $n_s$, the major complexity for a candidate pair is
\begin{equation}
\mathcal{O}(K_v K_t n_s), \qquad K_v \ll N_v,\ K_t \ll N_t.
\end{equation}
Because $K_v$ and $K_t$ are much smaller than the original patch/token lengths, the structured matching cost remains moderate and can be used as a lightweight reranking head after dual-encoder retrieval.

\subsection{Debiasing, Importance Weighting, and Prototype Consistency}

Background co-occurrence may create shortcut correlations in image-text matching. \method{} therefore introduces a background debiasing term that penalizes high similarity between low-response visual slots and the textual global feature:
\begin{equation}
\mathcal{L}_{debias} =
\frac{1}{B}\sum_{b=1}^{B}
\max(0,\cos(\mathbf{v}_{bg}^{(b)},\mathbf{g}_t^{(b)})-\delta),
\end{equation}
where $\mathbf{v}_{bg}$ is the aggregated low-response visual unit and $\delta$ is a margin.

Textual importance weighting further emphasizes entity, attribute, quantity, and relation phrases. For the $l$-th phrase unit, an importance gate is computed as
\begin{equation}
w_l^t=\sigma(\mathbf{w}_a^\top \mathbf{p}_l+b_a),
\qquad
\hat{\mathbf{p}}_l=w_l^t\mathbf{p}_l ,
\end{equation}
where $\sigma(\cdot)$ is the sigmoid function. The weighted phrase representation $\hat{\mathbf{p}}_l$ is used to construct the region-phrase cost matrix. The normalized weights can also define the textual marginal, $b_l=w_l^t/\sum_{l'}w_{l'}^t$, so that informative phrases receive a larger matching budget in the transport plan.

To improve cross-sample stability, a prototype consistency loss, inspired by prototypical representation learning and balanced clustering \cite{ref45,ref46}, is imposed over dynamic semantic units:
\begin{equation}
\mathcal{L}_{proto}
= - \frac{1}{M}\sum_{m=1}^{M}\sum_{r=1}^{R}
\tilde{z}_{mr}\log z_{mr}.
\end{equation}
Here $z_{mr}$ is the predicted prototype distribution and $\tilde{z}_{mr}$ is a balanced stop-gradient target.

\subsection{Training Objective and Inference}

The full objective is
\begin{equation}
\mathcal{L}=
\mathcal{L}_{con}
+\lambda_g \mathcal{L}_{gran}
+\lambda_b \mathcal{L}_{debias}
+\lambda_p \mathcal{L}_{proto}
+\lambda_c \mathcal{L}_{cons},
\end{equation}
where $\mathcal{L}_{con}$ is the bidirectional image-text contrastive loss, and $\mathcal{L}_{cons}$ maintains consistency between local semantic units and global representations.

During inference, the frozen dual encoder first retrieves a candidate set using the global similarity $S_g$. \method{} then applies dynamic granularity discovery and \ot{}-based region-phrase matching only to the candidate pairs, and ranks them by $S(I,T)$. This design keeps the indexable dual-encoder retrieval pipeline while adding fine-grained local evidence at the reranking stage.

\section{Experiments}

\subsection{Experimental Setup}

Experiments are conducted on Flickr30K and MS-COCO \cite{ref47,ref48}. Flickr30K follows the 1K test protocol. MS-COCO is evaluated under both the 5K full test setting and the 1K five-fold average setting. MS-COCO contains diverse and sometimes semantically close captions, a property also discussed in extended caption similarity studies \cite{ref49}. We report image-to-text (I2T) and text-to-image (T2I) Recall@1, Recall@5, Recall@10, and \rsum, which is the sum of the six recall metrics.

Unless otherwise specified, the backbone is OpenAI CLIP ViT-L/14 and is kept frozen. Only the lightweight \method{} head is trained. The slot number denotes the number of dynamic semantic units used in DGD. For MS-COCO, the default setting uses 12 visual/text slots, 512 hidden dimensions, 48 prototypes, 15 Sinkhorn iterations, and an \ot{} entropy coefficient of 0.07. For Flickr30K, it uses 8 visual/text slots, 256 hidden dimensions, 32 prototypes, 10 Sinkhorn iterations, and an \ot{} entropy coefficient of 0.10. The experiments are conducted on an NVIDIA A100 GPU.

\subsection{Main Retrieval Results}

\begin{table}[!t]
\centering
\caption{Main retrieval results on Flickr30K-1K and MS-COCO.}
\label{tab:main_results}
\tabfont
\setlength{\tabcolsep}{3.2pt}
\renewcommand{\arraystretch}{0.92}
\setlength{\tabcolsep}{0.2pt}\begin{tabular*}{\textwidth}{@{\extracolsep{\fill}}llrrrrrrr@{}}
\toprule
Benchmark & Method & I2T \R{1} & \R{5} & \R{10} & T2I \R{1} & \R{5} & \R{10} & \rsum{} \\
\midrule
\multirow{11}{*}{Flickr30K-1K}
& VSE++ (2018) \cite{ref50} & 52.9 & 80.5 & 87.2 & 39.6 & 70.1 & 79.5 & 409.8 \\
& SCAN (2018) \cite{ref14} & 67.4 & 90.3 & 95.8 & 48.6 & 77.7 & 85.2 & 465.0 \\
& VSRN (2019) \cite{ref15} & 71.3 & 90.6 & 96.0 & 54.7 & 81.8 & 88.2 & 482.6 \\
& IMRAM (2020) \cite{ref51} & 74.1 & 93.0 & 96.6 & 53.9 & 79.4 & 87.2 & 484.2 \\
& NAAF (2022) \cite{ref19} & 81.9 & 96.1 & 98.3 & 61.0 & 85.3 & 90.6 & 513.2 \\
& CHAN (2023) \cite{ref20} & 80.6 & 96.1 & 97.8 & 63.9 & 87.5 & 92.6 & 518.5 \\
& HREM (2023) \cite{ref52} & 83.3 & 96.0 & 98.1 & 63.5 & 87.1 & 92.4 & 520.4 \\
& LAPS (2024) \cite{ref23} & 85.1 & 97.7 & 99.2 & 74.0 & 93.0 & 96.3 & 545.3 \\
& EISIN (2025) \cite{ref53} & 83.3 & 96.2 & 98.3 & 63.6 & 87.3 & 92.0 & 520.7 \\
& Full \method{} (ours) & 86.6 & 98.3 & 99.4 & 67.62 & 89.28 & 94.58 & 535.78 \\
& \method-Fusion-Calib (ours) & \textbf{87.0} & 98.0 & 99.4 & 72.20 & 91.14 & 95.18 & 542.92 \\
\midrule
\multirow{11}{*}{MS-COCO 1K}
& VSE++ (2018) \cite{ref50} & 64.6 & 90.0 & 95.7 & 52.0 & 84.3 & 92.0 & 478.6 \\
& SCAN (2018) \cite{ref14} & 72.7 & 94.8 & 98.4 & 58.8 & 88.4 & 94.8 & 507.9 \\
& VSRN (2019) \cite{ref15} & 76.2 & 94.8 & 98.2 & 62.8 & 89.7 & 95.1 & 516.8 \\
& IMRAM (2020) \cite{ref51} & 76.7 & 95.6 & 98.5 & 61.7 & 89.1 & 95.0 & 516.6 \\
& NAAF (2022) \cite{ref19} & 80.5 & 96.5 & 98.8 & 64.1 & 90.7 & 96.5 & 527.2 \\
& CHAN (2023) \cite{ref20} & 81.4 & 96.9 & 98.9 & 66.5 & 92.1 & 96.7 & 532.6 \\
& HREM (2023) \cite{ref52} & 81.1 & 96.6 & 98.9 & 66.1 & 91.6 & 96.5 & 530.7 \\
& LAPS (2024) \cite{ref23} & 84.1 & 97.4 & 99.2 & 72.1 & 93.9 & 97.4 & 544.1 \\
& EISIN (2025) \cite{ref53} & 79.3 & 95.9 & 98.5 & 63.9 & 90.8 & 95.8 & 524.1 \\
& Full \method{} (ours) & 81.38 & 96.58 & 98.50 & 62.56 & 88.04 & 95.89 & 522.95 \\
& \method-Fusion-Calib (ours) & 81.72 & 97.50 & 98.82 & 66.18 & 91.52 & 96.88 & 532.62 \\
\midrule
\multirow{11}{*}{MS-COCO 5K}
& VSE++ (2018) \cite{ref50} & 41.3 & 71.1 & 81.2 & 30.3 & 59.4 & 72.4 & 355.7 \\
& SCAN (2018) \cite{ref14} & 50.4 & 82.2 & 90.0 & 38.6 & 69.3 & 80.4 & 410.9 \\
& VSRN (2019) \cite{ref15} & 53.0 & 81.1 & 89.4 & 40.5 & 70.6 & 81.1 & 415.7 \\
& IMRAM (2020) \cite{ref51} & 53.7 & 83.2 & 91.0 & 39.7 & 69.1 & 79.8 & 416.5 \\
& NAAF (2022) \cite{ref19} & 58.9 & 85.2 & 92.0 & 42.5 & 70.9 & 81.4 & 430.9 \\
& CHAN (2023) \cite{ref20} & 59.8 & 87.2 & 93.3 & 44.9 & 74.5 & 84.2 & 443.9 \\
& HREM (2023) \cite{ref52} & 62.3 & 87.6 & 93.4 & 43.9 & 73.6 & 83.3 & 444.1 \\
& LAPS (2024) \cite{ref23} & 67.1 & 88.6 & 94.3 & 53.0 & 79.5 & 87.6 & 470.1 \\
& EISIN (2025) \cite{ref53} & 59.6 & 84.4 & 91.2 & 42.0 & 72.3 & 82.5 & 432.1 \\
& Full \method{} (ours) & 64.78 & 85.76 & 91.98 & 46.93 & 70.72 & 79.42 & 439.59 \\
& \method-Fusion-Calib (ours) & 64.70 & 85.52 & 90.94 & 47.86 & 72.37 & 80.88 & 442.27 \\
\bottomrule
\end{tabular*}
\end{table}

Table~\ref{tab:main_results} shows that \method{} achieves competitive retrieval performance with a frozen CLIP backbone and a lightweight matching head. On Flickr30K-1K, Full \method{} obtains an \rsum{} of 535.78, while \method-Fusion-Calib reaches 542.92. This result is clearly higher than classical and recent baselines such as VSE++, SCAN, VSRN, IMRAM, NAAF, CHAN, HREM, and EISIN, and is close to LAPS. On MS-COCO 1K, \method-Fusion-Calib reaches an \rsum{} of 532.62, comparable to CHAN and higher than HREM and EISIN. On the larger MS-COCO 5K split, it achieves 442.27, higher than NAAF and EISIN and close to CHAN and HREM.

The advantage of \method{} is particularly reflected in top-ranked retrieval. For Flickr30K-1K, \method-Fusion-Calib achieves 87.0 I2T \R{1}, exceeding LAPS at 85.1. This indicates that dynamic semantic units and \ot{}-based matching are effective for determining the most plausible caption when several candidates are globally similar. The reason is not merely an increase in local interaction, but a change in the matching objects and the matching constraint. \dgd{} reduces fragmented patch/token noise by constructing compact region-phrase units, while \ot{} prevents local phrases from independently selecting the same visually salient region. Such structured local evidence is especially useful for \R{1}, where only the first-ranked candidate matters.

\subsection{Ablation Study}

\begin{table}[!htbp]
\centering
\caption{Core module ablation results.}
\label{tab:ablation}
\centering
\tabfont
\setlength{\tabcolsep}{2.0pt}
\renewcommand{\arraystretch}{0.92}
\begin{tabular}{llrrrr}
\toprule
Dataset & Variant & I2T \R{1} & T2I \R{1} & \rsum{} & $\Delta$\rsum{} \\
\midrule
COCO 5K & CLIP baseline & 56.34 & 36.50 & 390.97 & 0.00 \\
COCO 5K & \dgd{} only & 59.29 & 40.15 & 407.99 & +17.02 \\
COCO 5K & + \ot{} matching & 62.25 & 43.80 & 425.00 & +34.03 \\
COCO 5K & \makecell[l]{+ background debiasing\\+ textual importance weighting} & 63.77 & 45.68 & 433.74 & +42.77 \\
COCO 5K & + prototype consistency (Full \method) & 64.78 & 46.93 & 439.59 & +48.62 \\
\midrule
COCO 1K & CLIP baseline & 74.22 & 55.44 & 493.13 & 0.00 \\
COCO 1K & \dgd{} only & 76.73 & 57.93 & 503.57 & +10.44 \\
COCO 1K & + \ot{} matching & 79.23 & 60.42 & 514.00 & +20.87 \\
COCO 1K & \makecell[l]{+ background debiasing\\+ textual importance weighting} & 80.52 & 61.71 & 519.37 & +26.24 \\
COCO 1K & + prototype consistency (Full \method) & 81.38 & 62.56 & 522.95 & +29.82 \\
\midrule
Flickr30K & CLIP baseline & 85.20 & 64.92 & 525.94 & 0.00 \\
Flickr30K & \dgd{} only & 85.69 & 65.86 & 529.38 & +3.44 \\
Flickr30K & + \ot{} matching & 86.18 & 66.81 & 532.83 & +6.89 \\
Flickr30K & \makecell[l]{+ background debiasing\\+ textual importance weighting} & 86.43 & 67.30 & 534.60 & +8.66 \\
Flickr30K & + prototype consistency (Full \method) & 86.60 & 67.62 & 535.78 & +9.84 \\
\bottomrule
\end{tabular}
\end{table}

Table~\ref{tab:ablation} verifies the contributions of the main components. On COCO 5K, \dgd{} alone improves \rsum{} from 390.97 to 407.99, indicating that adaptive semantic units already provide useful local evidence. Adding \ot{} matching further increases \rsum{} to 425.00, showing that structured region-phrase matching is more effective than merely constructing local units. Debiasing with textual importance weighting and prototype consistency continue to improve the result to 433.74 and 439.59. Similar trends can be observed on COCO 1K and Flickr30K, which confirms the complementarity between semantic unit discovery and structured matching.

\subsection{Comparison of Fine-Grained Matching Mechanisms}

\begin{table}[!htbp]
\centering
\caption{Candidate promotion success rates (\%) of different fine-grained matching mechanisms under the Top-128 candidate pool.}
\label{tab:mechanism}
\tabfont
\setlength{\tabcolsep}{2.0pt}
\renewcommand{\arraystretch}{0.96}
\begin{tabular}{llrrrr}
\toprule
Dataset & Mechanism & Top10 & Top5 & Top1 & Gain \\
\midrule
Flickr30K & Max pooling & 32.99 & 26.29 & 13.57 & 0.00 \\
Flickr30K & Smooth Chamfer & 36.62 & 27.70 & 14.85 & +3.63 \\
Flickr30K & Hard assignment & 39.74 & 28.48 & 15.75 & +6.75 \\
Flickr30K & \ot{} matching & \textbf{43.64} & \textbf{31.77} & \textbf{17.99} & \textbf{+10.65} \\
\midrule
MS-COCO 5K & Max pooling & 25.30 & 20.22 & 12.22 & 0.00 \\
MS-COCO 5K & Smooth Chamfer & 27.93 & 21.76 & 13.02 & +2.63 \\
MS-COCO 5K & Hard assignment & 30.08 & 23.70 & 14.19 & +4.78 \\
MS-COCO 5K & \ot{} matching & \textbf{33.71} & \textbf{26.54} & \textbf{15.39} & \textbf{+8.41} \\
\bottomrule
\end{tabular}
\end{table}

\begin{figure}[!htbp]
\centering
\includegraphics[width=\columnwidth]{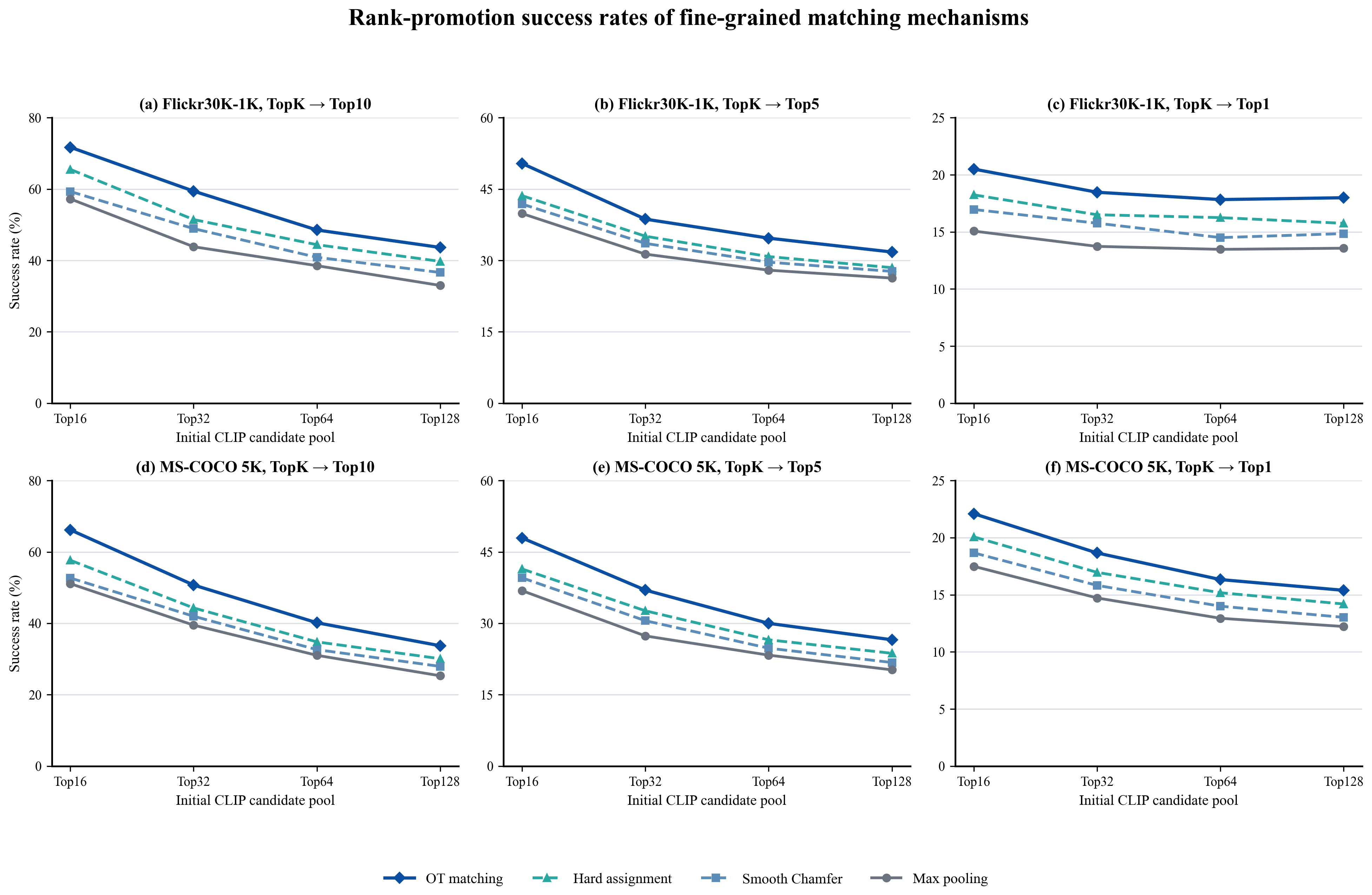}
\caption{Candidate promotion success-rate curves under different fine-grained matching mechanisms.}
\label{fig:mechanism}
\end{figure}

To directly examine whether a fine-grained mechanism can push correct candidates toward the top of the ranking list, we evaluate candidate promotion. A frozen CLIP global similarity first produces the initial ranking. Queries whose true matches are already in Top-128 but not yet in Top-10, Top-5, or Top-1 are selected. Different local matching mechanisms are then used to rerank the candidate set. Table~\ref{tab:mechanism} and Fig.~\ref{fig:mechanism} show that \ot{} matching consistently achieves the highest promotion success rate. The improvement is larger at Top-10 than at Top-1, which is expected because moving a candidate into the first position is more difficult and requires highly reliable global-local consistency.

\subsection{Parameter Sensitivity and Inference Efficiency}

\begin{table}[!htbp]
\centering
\caption{Numerical summary of key hyper-parameter sensitivity.}
\label{tab:sensitivity}
\centering
\tabfont
\setlength{\tabcolsep}{2.2pt}
\renewcommand{\arraystretch}{1.02}
\begin{tabular}{>{\raggedright\arraybackslash}p{0.16\columnwidth}>{\raggedright\arraybackslash}p{0.16\columnwidth}c>{\raggedright\arraybackslash}p{0.20\columnwidth}>{\centering\arraybackslash}p{0.18\columnwidth}>{\centering\arraybackslash}p{0.16\columnwidth}}
\toprule
Parameter & Search range & Points & Recommended value & Best \rsum{} & Endpoint drop \\
\midrule
Slot number $K$ & 1--48 & 14 & Flickr 8, COCO 12 & 535.78 / 522.95 / 439.59 & 67.19 / 85.58 \\
Prototype number $P$ & 4--768 & 14 & Flickr 32, COCO 48 & 535.78 / 522.95 / 439.59 & 47.49 / 117.75 \\
\ot{} coefficient $\varepsilon$ & 0.001--1.00 & 14 & Flickr 0.10, COCO 0.07 & 535.78 / 522.95 / 439.59 & 112.55 / 105.68 \\
Sinkhorn steps $n_s$ & 1--50 & 14 & 10--15 & 535.78 / 522.95 / 439.59 & 52.55 / 0.98 \\
Fusion weight $\beta$ & 0.00--1.50 & 16 & 0.10--0.20 & 535.78 / 522.95 / 439.59 & 48.62 / 192.75 \\
Rerank size $K_r$ & 4--1024 & 15 & 128--256 & 535.78 / 522.95 / 439.59 & 52.92 / 7.52 \\
\bottomrule
\end{tabular}
\end{table}

\begin{table}[!htbp]
\centering
\caption{Inference efficiency of \method.}
\label{tab:efficiency}
\tabfont
\setlength{\tabcolsep}{2.0pt}
\renewcommand{\arraystretch}{0.96}
\begin{tabular}{lrrrr}
\toprule
Stage & Candidates & Params (M) & Time (ms) & Memory (MB) \\
\midrule
CLIP encoding & 0 & 427.62 & 97.01 / 73.12 & 2299.90 \\
\method{} head & 1 & 1.35 & 0.097 & 2299.90 \\
\ot{} solver & 1 & 0.00 & 0.049 & -- \\
Top-64 rerank & 64 & 1.35 & 6.24 & 2299.90 \\
Top-128 rerank & 128 & 1.35 & 12.47 & 2299.90 \\
Top-256 rerank & 256 & 1.35 & 24.94 & 2299.90 \\
\bottomrule
\end{tabular}
\end{table}

Table~\ref{tab:sensitivity} summarizes the sensitivity of six key parameters. Here $K$ denotes the number of dynamic semantic slots rather than a manually defined scale. It has a clear effective range: too few slots compress multiple entities and relations into a small number of units, while too many slots introduce redundancy. The prototype number controls the capacity of cross-sample semantic priors. The \ot{} entropy coefficient shows a middle-range optimum: a too small value makes the transport plan close to hard matching, while a too large value over-smooths the alignment. Sinkhorn iteration becomes stable after about 10--15 steps. The fusion weight $\beta$ and the reranking size $K_r$ reflect how \method{} interacts with dual-encoder retrieval. A moderate $\beta$ and a candidate range of 128--256 balance ranking gain and computational cost.

Table~\ref{tab:efficiency} shows that the additional parameters of the \method{} head are only 1.35M. The per-candidate overhead of the \ot{} solver is small. Even for Top-128 reranking, the extra inference time is about 12.47 ms under the tested setting. Therefore, \method{} can act as a lightweight fine-grained ranking module after global retrieval.

\begin{figure}[!htbp]
\centering
\includegraphics[width=\columnwidth]{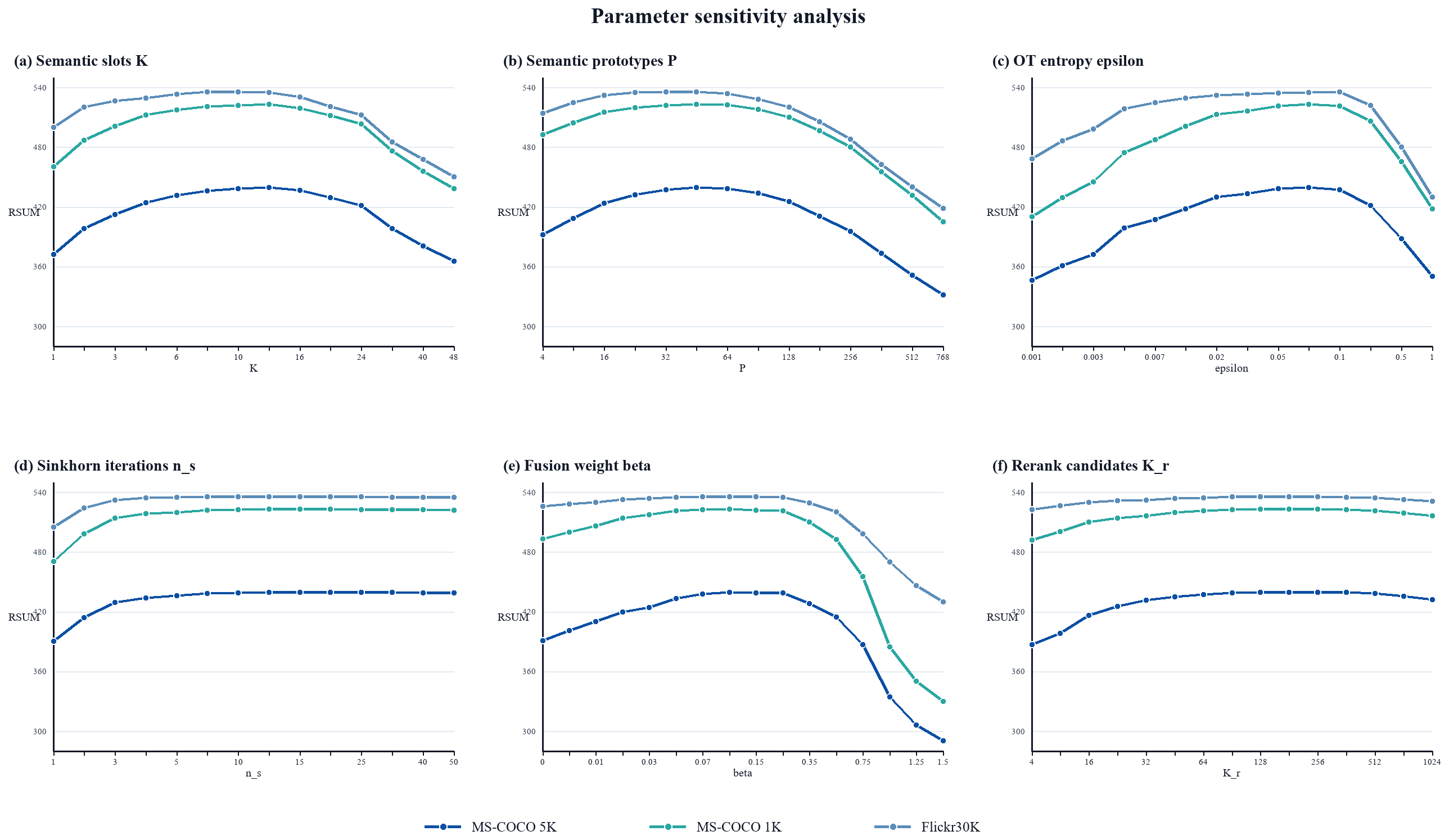}
\caption{Key parameter sensitivity analysis.}
\label{fig:sensitivity}
\end{figure}

\subsection{Cross-Dataset Generalization}

\begin{table}[!t]
\centering
\caption{Cross-dataset generalization analysis.}
\label{tab:cross}
\centering
\tabfont
\setlength{\tabcolsep}{3.2pt}
\renewcommand{\arraystretch}{0.92}
\fontsize{6}{7.5}\selectfont\setlength{\tabcolsep}{0.7pt}\begin{tabular*}{\textwidth}{@{\extracolsep{\fill}}lllrrrrrrr@{}}
\toprule
Training set & Test set & Method & I2T \R{1} & \R{5} & \R{10} & T2I \R{1} & \R{5} & \R{10} & \rsum{} \\
\midrule
MS-COCO 5K & Flickr30K & CLIP baseline & 85.10 & 97.40 & 99.20 & 64.90 & 87.28 & 92.00 & 525.88 \\
MS-COCO 5K & Flickr30K & Full \method{} & 86.40 & 98.30 & 98.70 & 68.04 & 88.66 & 91.88 & 531.98 \\
MS-COCO 5K & Flickr30K & \method-Fusion-Calib & 88.70 & 98.20 & 98.60 & 73.78 & 90.60 & 92.56 & 542.44 \\
\midrule
Flickr30K & MS-COCO 5K & CLIP baseline & 56.36 & 79.26 & 86.58 & 36.51 & 61.02 & 71.13 & 390.86 \\
Flickr30K & MS-COCO 5K & Full \method{} & 56.64 & 79.34 & 87.00 & 37.06 & 61.63 & 71.62 & 393.29 \\
Flickr30K & MS-COCO 5K & \method-Fusion-Calib & 58.58 & 81.26 & 88.12 & 43.12 & 67.36 & 76.00 & 414.44 \\
\bottomrule
\end{tabular*}
\end{table}

Table~\ref{tab:cross} reports cross-dataset transfer results. Training on MS-COCO and testing on Flickr30K brings clear improvements over the CLIP baseline, indicating that the learned region-phrase alignment can transfer to a smaller retrieval benchmark. The reverse direction, from Flickr30K to MS-COCO 5K, is more challenging because the test collection is larger and contains more diverse image-text descriptions. Even in this setting, \method-Fusion-Calib still improves \rsum{} from 390.86 to 414.44. These results suggest that \method{} has a certain level of cross-dataset adaptability, while its effect is influenced by differences between training and test distributions.

\subsection{Visualization Analysis}

\begin{figure}[!htbp]
\centering
\includegraphics[width=\columnwidth]{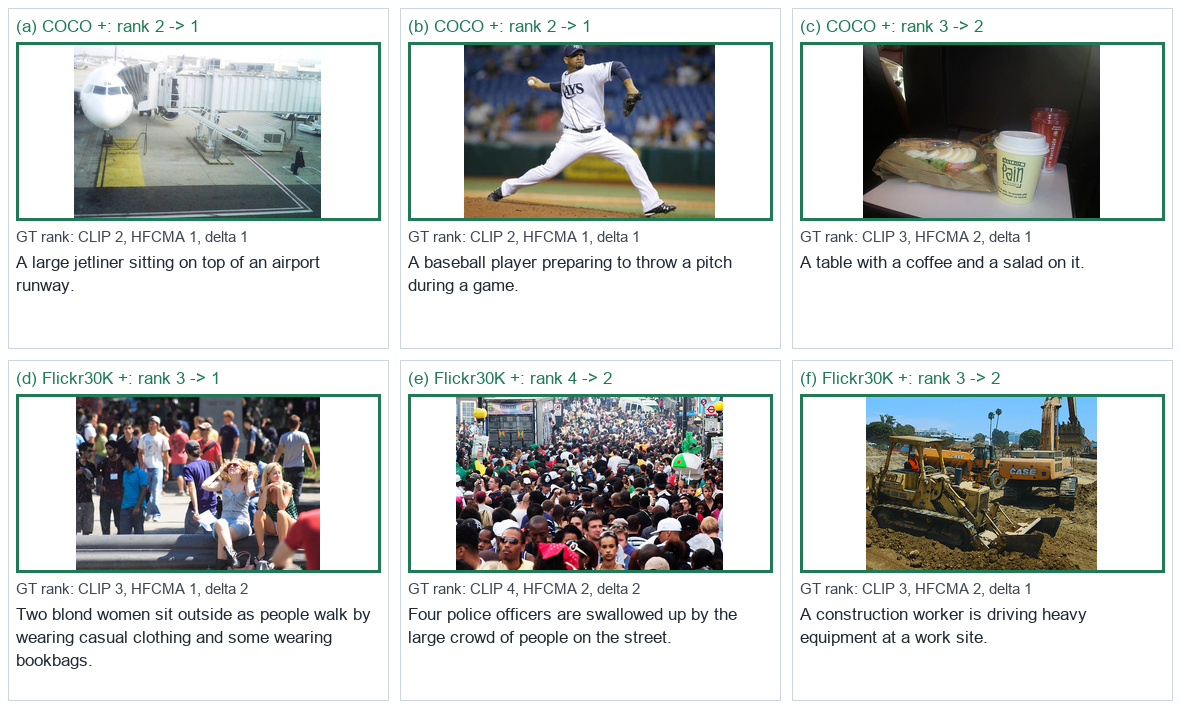}
\caption{Visualization of ranking-improvement cases.}
\label{fig:success_vis}
\end{figure}

Visualization is used to inspect whether \method{} provides interpretable local evidence. Fig.~\ref{fig:success_vis} presents representative ranking-improvement cases. The corrected candidates are visually close to the original CLIP top results, but differ in attributes, quantities, local objects, or spatial relations. The dynamic semantic units make these discriminative fragments explicit, and the \ot{} transport plan assigns them under a global matching constraint. This explains why \method{} can move some correct candidates from the candidate pool to the top positions.

\begin{figure}[!htbp]
\centering
\includegraphics[width=\columnwidth]{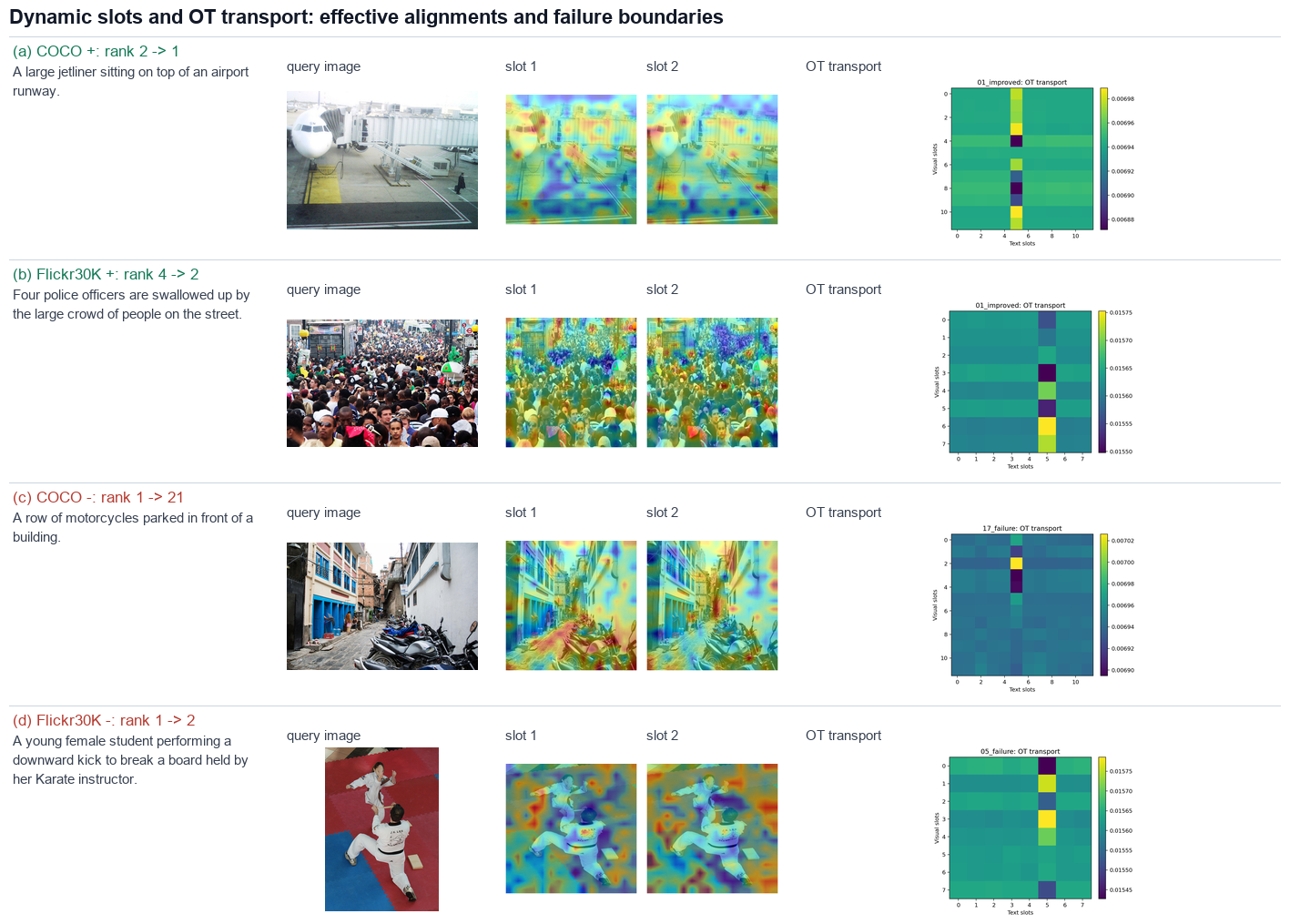}
\caption{Dynamic visual slots and local alignment evidence from the \ot{} transport matrix.}
\label{fig:slot_ot}
\end{figure}

Fig.~\ref{fig:slot_ot} shows the local alignment evidence behind the ranking changes. Active visual slots tend to cover semantically meaningful regions, while textual slots correspond to phrases involving objects, attributes, and relations. The transport matrix concentrates on plausible region-phrase pairs while retaining soft uncertainty, which is consistent with the role of \ot{} as a structured many-to-many matching mechanism.

\begin{figure}[!htbp]
\centering
\includegraphics[width=\columnwidth]{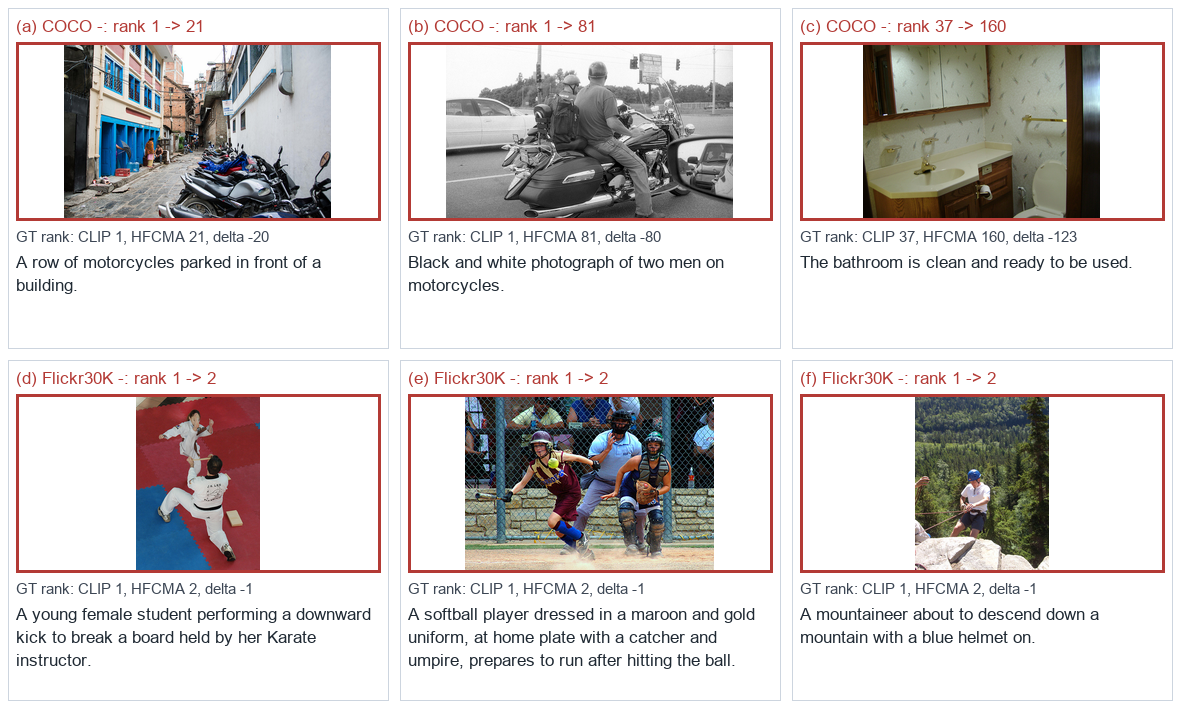}
\caption{Visualization of ranking-degradation and failure cases.}
\label{fig:failure_vis}
\end{figure}

Fig.~\ref{fig:failure_vis} further reports ranking-degradation and failure cases. These examples usually contain crowded scenes, overlapping objects, weakly localized attributes, or near-duplicate candidates. In such cases, dynamic slots may respond to background or visually salient but semantically irrelevant regions, and the transport plan may amplify this imperfect local evidence. The failure analysis indicates that \method{} is most reliable when the caption contains local cues that can be grounded by stable region-phrase correspondences.

\section{Discussion}

The experimental results indicate that the core issue in fine-grained image-text retrieval is not only how to aggregate local similarities, but also what local units should be matched. Fixed patches and tokens are computationally convenient, but they are not always consistent with semantic units. Dynamic granularity discovery organizes local features into compact visual regions and textual phrases, allowing subsequent matching to operate on units that better represent entities, attributes, and relations. \ot{} matching further places these units under a unified transport constraint, avoiding independent local maximum selection. In this sense, the hierarchical nature of HFCMA lies in the progressive transition from low-level patch/token features through adaptive region-phrase semantic units to structured cross-modal matching combined with global retrieval scoring.

The ablation results provide the most direct structural evidence: \dgd{} brings stable improvement, \ot{} strengthens structured matching, and auxiliary regularization further improves stability. Fusion and calibration are useful in practical retrieval because local fine-grained scores should cooperate with global semantic anchors rather than replace them. This point is important for large-scale datasets such as MS-COCO, where many candidates are visually or semantically close.

Several limitations remain. First, dynamic slots are not explicit detection boxes or segmentation masks, and semantic drift may occur in crowded scenes, occlusion, or near-duplicate objects. Second, calibration depends on the candidate distribution and should be fixed by validation protocols rather than repeatedly tuned on the test set. Third, different CLIP backbones may influence the gain of the fine-grained head. Finally, current experiments focus on English short-text retrieval benchmarks, whose brittleness and evaluation sensitivity have also been noted in recent benchmark analyses \cite{ref54}. Future work may extend the framework to multilingual descriptions, long-form narratives, and noisy open-domain image-text collections.

\section{Conclusion}

This paper proposes \method{}, a hierarchical fine-grained cross-modal alignment framework for image-text retrieval. The method adaptively constructs visual regions and textual phrases through dynamic granularity discovery, and solves a globally constrained region-phrase transport matrix with entropy-regularized optimal transport. Compared with methods that rely on fixed patches/tokens and heuristic local aggregation, \method{} improves both the matching units and the matching mechanism.

Experiments on Flickr30K and MS-COCO show that \method{} can enhance retrieval performance under a frozen CLIP dual-encoder framework with only a lightweight additional head. Ablation, matching mechanism comparison, parameter sensitivity, efficiency analysis, cross-dataset generalization, and visualization further verify the complementary role of dynamic semantic unit construction and \ot{}-based structured matching. The main conclusion is that adaptive semantic unit construction combined with structured set matching provides a feasible path for improving fine-grained discrimination in image-text retrieval.

\section*{Acknowledgements}
This work was supported by the Zhejiang Provincial Natural Science Foundation of China (Grant No. LY23G020007).
\section*{Competing Interests}
The authors declare that they have no competing interests.
\section*{Data and Code Availability}
The Flickr30K and MS-COCO datasets used in this study are publicly available from their respective providers. The HFCMA code is available at \url{https://github.com/sungkites/HFCMA}.

\end{document}